# Random vector generation of a semantic space

**Jean-François Delpech** **Sabine Ploux**
Institut des Sciences Cognitives
UMR5304 CNRS - Université de Lyon
67, boulevard Pinel
69675 BRON cedex, France
jfdelpech@gmail.com sploux@isc.cnrs.fr

We show how random vectors and random projection can be implemented in the usual vector space model to construct a Euclidean semantic space from a French synonym dictionary. We evaluate theoretically the resulting noise and show the experimental distribution of the similarities of terms in a neighborhood according to the choice of parameters. We also show that the Schmidt orthogonalization process is applicable and can be used to separate homonyms with distinct semantic meanings. Neighboring terms are easily arranged into semantically significant clusters which are well suited to the generation of realistic lists of synonyms and to such applications as word selection for automatic text generation. This process, applicable to any language, can easily be extended to collocations, is extremely fast and can be updated in real time, whenever new synonyms are proposed.

## 1. Introduction

In their seminal work, Ploux and Victorri [1] have used synonymy relations deduced from French electronic dictionaries to create semantic spaces around French words and their neighbors. Their definition of “synonymy” is fairly broad and includes hyponymy (*moineau* and *oiseau*), hyperonymy (*arme* and *pistolet*) or even non-synonymous, but related terms (*autocar* and *automobile*); however, in their work, true synonyms (i.e. terms which are more or less interchangeable) form cliques of the graph of synonyms, i.e. maximally complete subgraphs. While this is very interesting from a theoretical standpoint, as it then becomes straightforward to evaluate an interclique distance (or degree of separation) between any two terms in the graph (as long as neither belongs to an island, such as *lapereau* and *lapinot*), it is not very useful in practice. For example, an author in search of the right term may well not be interested in strict synonyms; terms with related or even opposed meanings can often be preferable in rhetorical figures. Also, in many applications such as automatic text generation, a well-defined and mathematically well behaved semantic distance between terms is often a prerequisite.

In this report, we show how an Euclidean semantic distance can quickly and easily be constructed from Ploux and Victorri's database (which contains 54,685 terms and 116,694 cliques).

## 2. Construction of a semantic space

### *2.1. The vector space model*

Since the pioneering work of Salton [2, 3], it is well understood that any combination of terms, such as a clique, can be seen as a vector in a space where each dimension represents a distinct term (or lemma.)

$$\mathcal{C}_j = (t_{1,j}, t_{2,j}, \ldots, t_{t,j}) \tag{1}$$

This representation is extremely fruitful and forms the basis of numerous information retrieval systems; it suffers however from a severe limitation in that each term is orthogonal to each other. Of course, the dual equation from Equation 1,

$$\mathcal{T}_k = (c_{1,k}, c_{2,k}, \ldots, c_{s,k}) \tag{2}$$

may be used to compute term distances (or similarities) but the very high dimensionality of the subtending space makes such distances difficult to compute and to interpret: this is the “curse of dimensionality”.

### 2.2. Overlap similarity between terms

If $D_i$ with cardinality $d_i$ is the set of distinct terms occurring in all the cliques containing term $t_i$, we define the overlap similarity between two terms as the cardinality $d_{i,k}$ of the intersection $D_i \cap D_k$ (each word being counted only once.) Obviously, $d_{i,k} = 0$ for most $(i, k)$ pairs, since for any $i$ the total number of distinct terms in the database is much larger than $d_i$, which ranges from $2$ to $243$ in Ploux and Victorri's [3] database with an average value of 8.5.

### 2.3. Contexonyms

“Contexonyms” are words which co-occur in a given context (as for example in the same sentence of a corpus); while they are not synonyms, they are obviously closely related. Ji, Ploux and Wehrli [4] have proposed an automatic contexonym organizing model (ACOM) which relies on counting of co-occurrences and evaluating their probabilities to automatically produce and organize contexonyms for a target word. The test results, after training on an English corpus maintained by Project Gutenberg, show that the model is able to classify contexonyms as well as to reflect words' minute usage and nuance.

### 2.4. Latent semantic indexing

Dimensionality reduction can be achieved by a low-rank approximation of the term-document matrix. This can be done by Latent Semantic Indexing [5], which reduces dimensionality through a singular value decomposition (SVD) of the term-document matrix, retaining only a comparatively small number of the largest singular values. This method has been very successfully used for document indexing and retrieval. It suffers nevertheless from limitations:

- SVD is computationally intensive, even though the large term-document matrix is very sparse;
- There is no really satisfactory way to increment the results as new terms/documents become available.

More importantly, it is not well suited to generating a semantic space from cliques. The resulting, lower dimensional space is the best approximation, in the least squares sense, of the position of any term belonging to the whole set of cliques: the distance *between any pair of terms* will be optimal, while what is really of interest from the present perspective is the accurate determination of distances *between semantic neighbors*.

As a test, a SVD decomposition of the clique-term matrix (of which eq. 2 is a row) was performed. It reduced the matrix size from 54,685 x 116,694 to 54,685 x 250, meaning that each term was associated with a vector having 250 orthogonal components. The decomposition, which took 134 sec. on a desktop computer, was clearly unsatisfactory as the singular values decayed very slowly, from 63.31 for the first coordinate to 37.4 for the 250$^{th}$ one. According to this computation, the first few neighbors of *rapsode* would be *rimailleur, rimeur, versificateur, métromane, fils d'Apollon, favori des Muses, favori du Parnasse, nourrisson des Muses, héros du Pinde, mâche-laurier*, all with similarities extremely close to 1.0. While clearly in the right neighborhood, this seems to be of limited usefulness; note however that in practice a restriction to the first two or three largest singular values may often yield useful information [6].

### 2.5. Neural networks

Word order is not considered in this publication, but it should be mentioned for completeness that neural networks are often used in natural language processing to encode word sequences (see [7] for an extended review and bibliography). In a recent publication, Mikolov *et al.* [8] have introduced two novel model architectures for computing continuous vector representations of words from very large data sets. They report large improvements in accuracy at a computational cost which is still substantial, but that they claim is much lower than previous architectures. An interesting consideration is that, according to Mikolov *et al.* [9], the learned vectors explicitly encode many linguistic regularities and patterns.

## 2.6. Random vectors and random projection

### 2.6.1. Remarks on high-dimensionality spaces

Dasgupta [10] points out that intuitions valid in a low-dimensionality space may be totally misleading in a high-dimensionality space. For example, a set of points picked at random from the unit ball

$$\{x \in V : \|x\| < 1\} \tag{3}$$

will have some significant fraction near the origin, say within distance $1/2$ if $d = 3$, but this fraction becomes rapidly vanishingly small as the dimension $d$ becomes large; for example for $d = 250$, $d^{250} = 0.5^{250} \approx 5.5 \times 10^{-76}$.

Another useful remark is that while obviously one cannot create more than $d$ orthogonal vectors in a space of dimension $d$, one can create an exponentially large number of vectors quasi-orthogonal to each other; in other words [10], a set of $\exp(\mathcal{O}(\epsilon^2 d))$ vectors picked at random will with high probability be quasi-orthogonal, i.e. have angles of $90 \pm \epsilon$ with each others. The seed vectors referred to below will be selected from such a set $\mathbb{S}_d$ .

While an orthogonal projection will in general reduce the average distance between points, it is also known, as shown by Johnson and Lindenstrauss in an often cited paper [11], that distances may be almost perfectly preserved for any $n$ points in an arbitrary number $d$ of dimensions when projected to a random subspace of $\mathcal{O}(\log n)$ dimension.

### 2.6.2. Building random vectors

The comparatively recent method of random projection [12], [13], [14] is based on these three preceding remarks and proceeds as follows:

1. Uniquely associate with each term $t_i$ a random seed vector $s_i \in \mathbb{S}_d$ having $d$ independent coordinates;
2. Associate with each clique $c_k$ the vector $\mathcal{C}_k = \sum_{i \in c_k} \rho_i^k s_i$ where $i \in c_k$ refers to the set of terms found in clique $c_k$ and $\rho_i^k$ is a function of the number of occurrences in $c_k$ of term $t_i$ and of the weights associated with $t_i$;
3. Finally, associate with each term $t_i$ the (suitably weighted) sum of the vectors of the cliques in which $t_i$ appears, $\mathcal{T}_i = \sum_{k \ni t_i} \mathcal{C}_k$ where $k \ni t_i$ refers to the cliques containing $t_i$. In what follows, we shall assume without loss of generality that the term vectors $\mathcal{T}_i$ are normalized to unity.

Obviously, each term vector $\mathcal{T}_i$ is now embedded in a $d$-dimensional Euclidean semantic space and the similarity $\sigma_{ij}$ between terms $t_i$ and $t_j$ is the scalar product of the associated term vectors:

$$\sigma_{ij} = \langle \mathcal{T}_i | \mathcal{T}_j \rangle \tag{4}$$

It is easy to see that $\sigma_{ij}$ ranges from -1 to 1. It is sometimes more convenient to consider the distance $\mathcal{D}_{ij}$ which is related to the similarity by $\mathcal{D}_{ij} = \sqrt{2(1 - \sigma_{ij})}$ and ranges from $0$ (same $t_i$ and $t_k$) to $2$ (exactly opposite terms; note however that owing to the extreme sparsity of a high-dimensional space, the neighborhood exactly opposite a term is in practice always empty.)

### 2.6.3. Locality property

Building a term vector $\mathcal{T}_i$ by the process described above involves only the terms pertaining to the set $D_i$ defined in section 2.2. It is thus a purely local process: updating the semantic space requires only a few ten or a few hundred operations, orders of magnitude less than its initial generation (provided small changes to the weights are neglected, which is usually acceptable as they are logarithmic in term frequency and inverse document frequency.)

This does not imply that the similarity of term $t_i$ with term $t_k \notin D_i$ is zero, even though $t_k \notin D_i \implies t_i \notin D_k$, since $t_i$ and $t_k$ may well have neighbors in common. For example *carotte* and *fraude* have a degree of separation of 2 but a similarity of 0.364.

The seed vectors $s_i$ are not quite orthogonal and the scalar product $\langle s_i | s_j \rangle$ will usually be small, but not zero. Thus, even for uncorrelated term vectors $\mathcal{T}_i$ and $\mathcal{T}_j$, their similarity $\sigma_{ij}$ will usually be small but non-zero. This induce an unavoidable noise which is studied below in some detail.

### *2.7. Practical implementation*

A normalized seed vector embedded in a $d$-dimensional space has $d$ coordinates of which $d - 2m$ are 0, $m$ are $+1/\sqrt{2m}$ and $m$ are $-1/\sqrt{2m}$. Having the same number of positive and negative coordinates ensures that the scalar product of two seed vectors is 0 on the average. As seed vectors need to be very close to orthogonal with each others, the number $2m$ of non-zero coefficients must be substantially smaller than the dimension $d$.

The number of available, distinct seed vectors is the product of the number of combinations $\binom{d}{2m}$ of $2m$ non-zero coordinates amongst $d$ coordinates, times the number of ways $\binom{2m}{m}$ of distributing $m$ positive and $m$ negative coordinates amongst these non-zero coordinates :

$$N_{seed}(d, m) = \binom{d}{2m} \times \binom{2m}{m} \tag{5}$$

In practice, $N_{seed}$ should be much larger than the number of distinct terms to guarantee a negligible collision probability (i.e. two distinct terms having the same seed vector). This condition is already amply met with $m \geq 4$ as $N_{seed}(250, 4) \approx 2.4 \times 10^{16}$ when a dimension $d = 250$ is selected.

Given $d$ and $m$ and noting for simplicity $p = 2m/d$, the probability $P_{overlap}(v, d)$ of an overlap $v$ between two randomly selected seed vectors is

$$P_{overlap}(v, d) = \binom{2m}{v} \times p^v \times (1-p)^{(2m-v)} \tag{6}$$

When two randomly selected, normalized seed vectors have an overlap of $v$ non-zero coordinates, their scalar products will be arranged symmetrically around zero and vary in discrete steps of $1/2m$. An overlap of $1$ will generate the two scalars $1/2m$ and $-1/2m$ with probabilities $1/2$, an overlap of 2 will generate $2/2m$ with probability $1/4$, 0 with probability $1/2$, and $-2/2m$ with probability $1/4$; more generally, an overlap $v$ will generate the scalar $s/2m$ with the probability

$$P_{scalar}(v, s) = \binom{v}{q_s} \times 2^{-v} \tag{7}$$

where the factors $q_s = (v + s)/2$ are restricted to integer values and by virtue of the identity $\sum_{k=-v}^{v} \binom{k}{q_k} \equiv 2^v$.

It can be seen from equations 6 and 7 that the noise decreases more or less linearly with $d$. Theoretical and experimental scalar products of two seed vectors, as computed from equations 6 and 7, are plotted in the next figure (next page) where:

- The light vertical lines are increments of $0.01\sigma$, the heavier vertical lines are at 0, $-0.1\sigma$ and $+0.1\sigma$.
- The two horizontal lines are at 1.0 and 0.136.
- The red dots are computed by taking the scalar products of 1,000,000 'term vectors' each synthesized by the addition of 5 random seed vectors. If instead we do the statistics directly on seed vectors, the result is unchanged except that the dots now occur only at multiples of 0.01 and the total is accordingly 5 times larger.
- The black dots are statistics over 40,000 points, starting with 10,000, taken from the tail of the neighbors of an arbitrary term (here *rapsode*.)
- The purple dots are a Gaussian with a standard deviation of $0.063$.

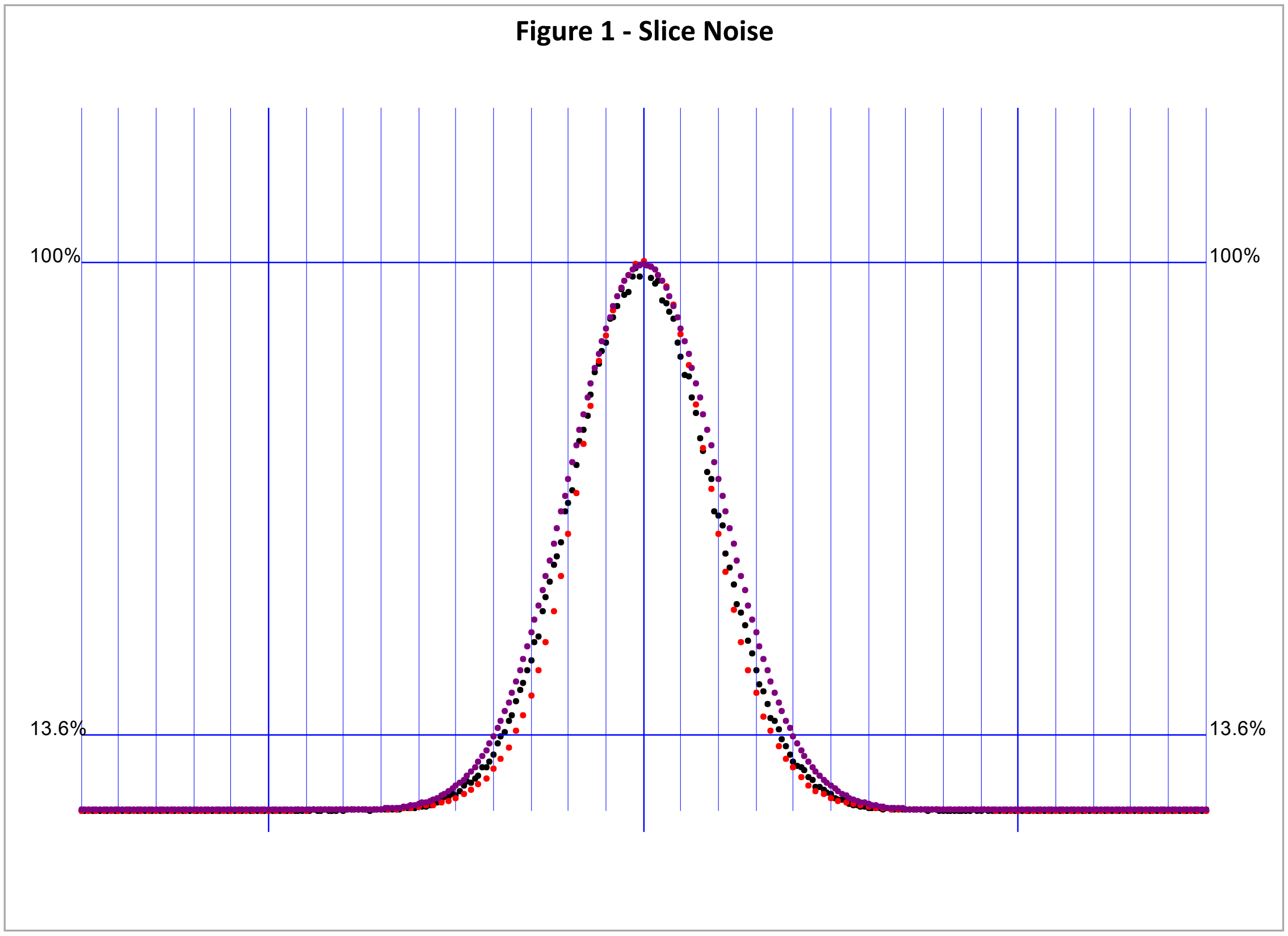

# 3. Results

Obviously, the size of the database of the term vectors $\mathcal{T}_i$ is itself linearly dependent on the dimension; for $d = 2500$, each vector occupies 10 kB if a single coordinate is represented by a 4-byte floating point number. A database of 1,000,000 distinct terms would thus occupy 10 GB with this elementary data structure; however, for many applications, $d = 250$ will be sufficient and/or more sophisticated data structures may be implemented. Computation times will also increase more or less linearly with $d$, because they mostly involve stepping through all the dimensions.

### *3.1. Term neighbors*

The following four figures have been constructed by compiling eight independent databases of $54,685$ term vectors for each of the four indicated $(m, d)$ couples; *tf-idf* statistical weights were used. Typically, on a small desktop computer, the compilation time is 3 to 4 seconds for $d = 250$ and 20 to 30 seconds for $d = 2500$.

Abscissas are proportional to the logarithm of the neighbor's rank (From 1 for *maison* in the upper right corner to 1000 in the lower left corner) and ordinates are the similarities to *maison*. For a given neighbor, the horizontally aligned red dots represent the eight scalars $\sigma_i$ computed from the eight databases and the thicker, black dot is the average value $\sigma_{avg}$ of the eight scalars. The neighbors are arranged in non-increasing order of their $\sigma_{avg}$ with *maison*.

Even though the diameter $d_{maison}$ of *maison* as defined in section 2.2 is only 98, there are several hundred significant neighbors: while most close neighbors belong to the set $D_{maison}$, many lie at more than one degree of separation from each others (their cliques are separated by more than one vertex). Also, it can be seen, as expected, that the noise is inversely proportional to $\sqrt{d}$ but not very dependent on $m$ (in fact, the only noticeable effect of a lower $m$ is that there are more outliers) and that a standard deviation of about $0.063$ is not unrealistic for $d = 2500$.

**Figure 2 - Neighbors of *maison***

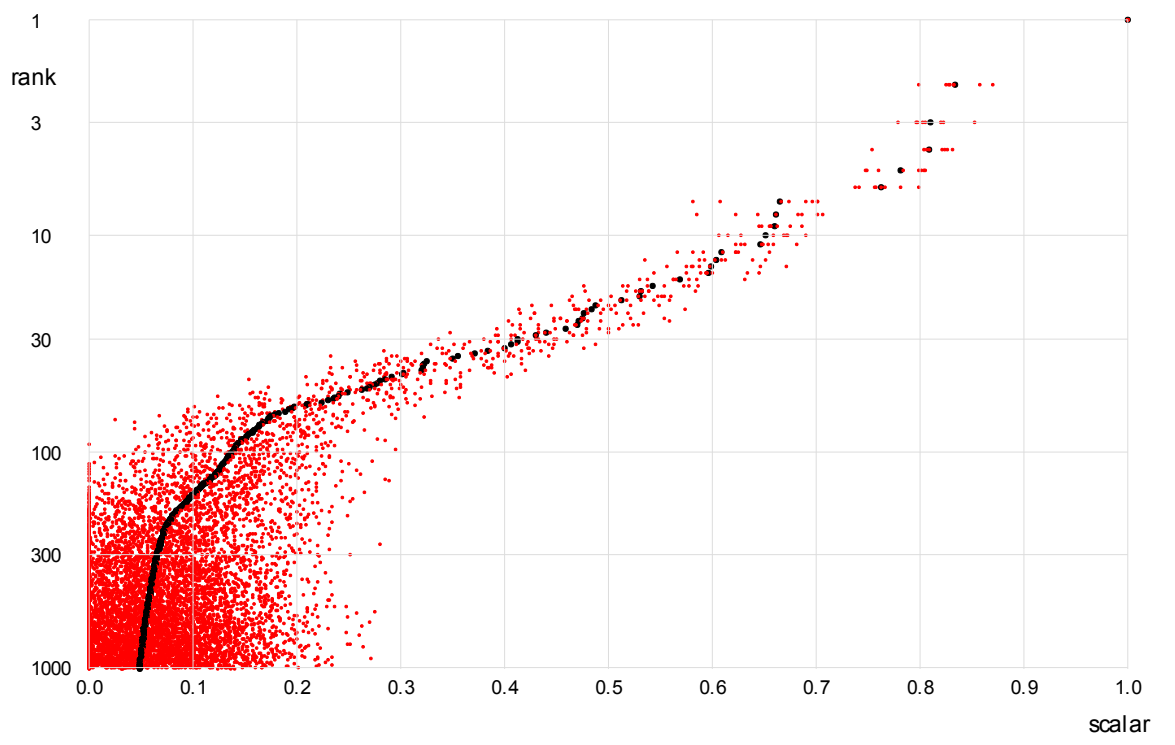

$m = 5 \qquad d = 250$

**Figure 3 - Neighbors of *maison***

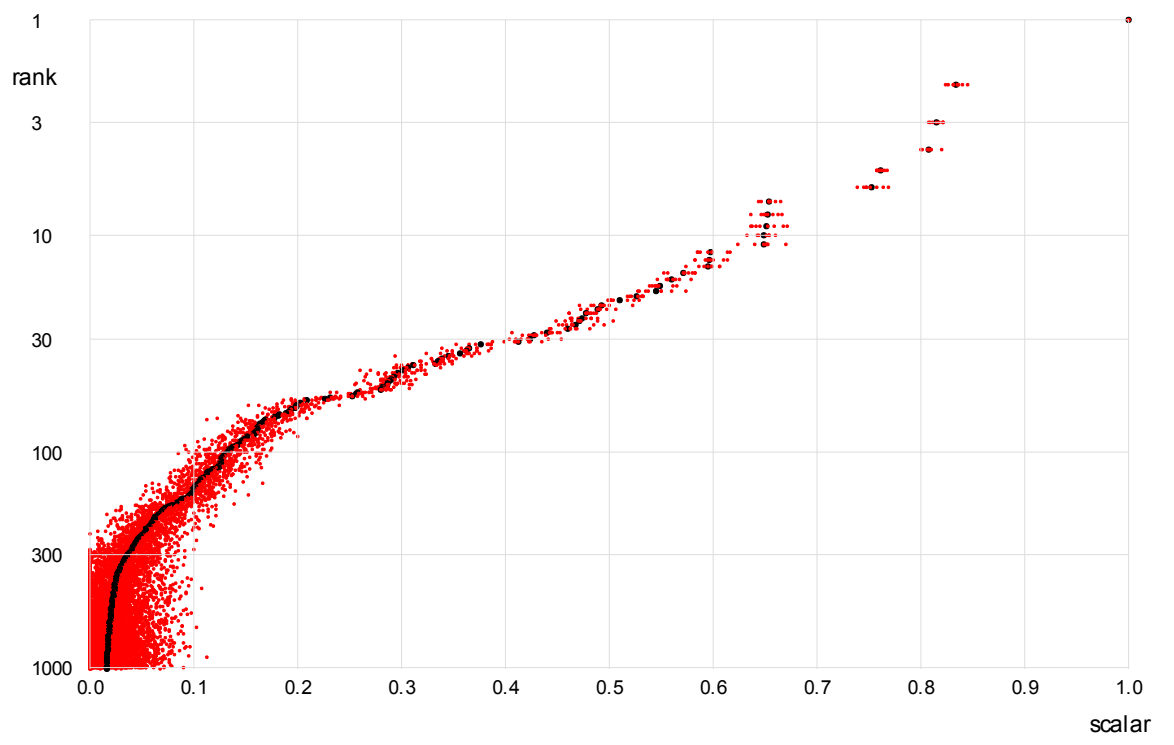

$m = 10 \qquad d = 2500$

**Figure 4 - Neighbors of *maison***

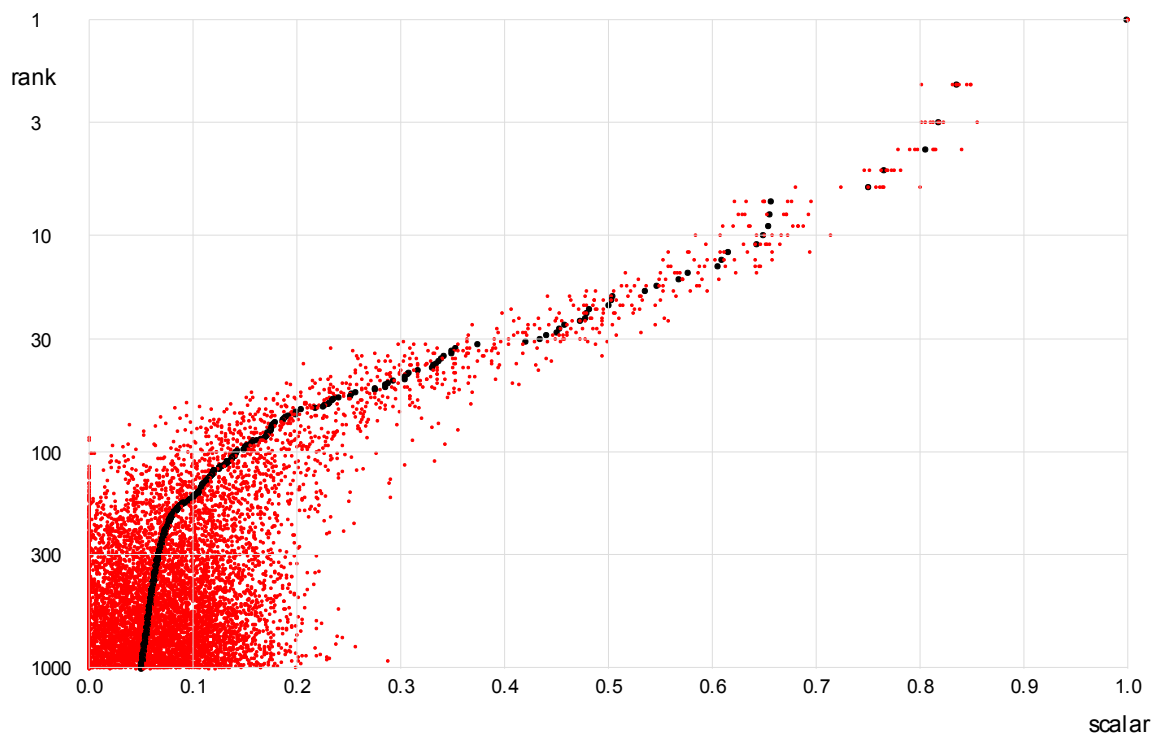

$m = 40 \qquad d = 250$

**Figure 5 - Neighbors of *maison***

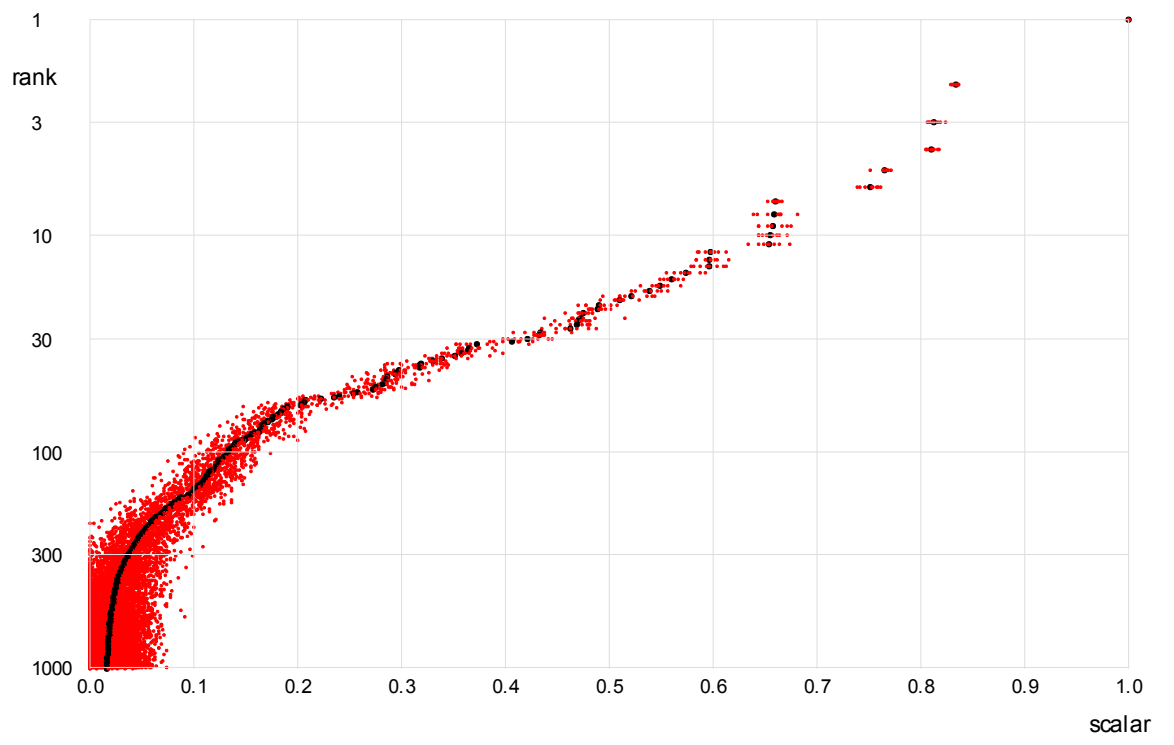

$m = 50 \qquad d = 2500$

In what follows, unless otherwise noted, we'll use $d = 2500$ and $m = 50$. The size of the file containing the 54,685 term vectors is then 547,724,964 bytes, including some overhead. The number of available seed vectors being $\approx 2.0 \times 10^{255}$ the risk of collision is totally negligible.

The 100 first neighbors of *maison* are listed by decreasing similarity in table 1 next page. It is clear that the proximity decreases with $\sigma$, but that those neighboring words are all reasonably close to *maison* in its various meanings.

**Table 1 - First 100 neighbors of *maison***

| From 1 to 20 | | | From 21 to 40 | | | From 41 to 60 | | | From 61 to 80 | | | From 81 to 100 | | |
|---|---|---|---|---|---|---|---|---|---|---|---|---|---|---|
| 1 | 1.000 | maison | 21 | 0.499 | chez-soi | 41 | 0.309 | ménage | 61 | 0.207 | ermitage | 81 | 0.163 | domestique |
| 2 | 0.843 | demeure | 22 | 0.498 | cassine | 42 | 0.307 | bâtiment | 62 | 0.204 | appartement | 82 | 0.162 | plaque_de_blindage |
| 3 | 0.820 | habitation | 23 | 0.491 | cabane | 43 | 0.306 | case | 63 | 0.198 | cagna | 83 | 0.159 | mas |
| 4 | 0.810 | logis | 24 | 0.486 | gourbi | 44 | 0.301 | taudis | 64 | 0.191 | reposée | 84 | 0.155 | tanière |
| 5 | 0.767 | domicile | 25 | 0.480 | gîte | 45 | 0.300 | chalet | 65 | 0.190 | chartreuse | 85 | 0.154 | cache |
| 6 | 0.762 | pénates | 26 | 0.477 | bercail | 46 | 0.292 | pavillon | 66 | 0.188 | domesticité | 86 | 0.152 | niche |
| 7 | 0.677 | home | 27 | 0.460 | masure | 47 | 0.292 | villa | 67 | 0.187 | garde-meubles | 87 | 0.151 | rendez-vous_de_chasse |
| 8 | 0.669 | mesnil | 28 | 0.436 | asile | 48 | 0.287 | chaumière | 68 | 0.186 | gabionnade | 88 | 0.151 | repaire |
| 9 | 0.661 | chacunière | 29 | 0.428 | château | 49 | 0.278 | manse | 69 | 0.182 | lignée | 89 | 0.149 | clinique |
| 10 | 0.659 | foyer | 30 | 0.404 | immeuble | 50 | 0.275 | hôtel_particulier | 70 | 0.180 | caponnière | 90 | 0.148 | gloriette |
| 11 | 0.653 | train_de_maison | 31 | 0.403 | hutte | 51 | 0.265 | isba | 71 | 0.172 | fermette | 91 | 0.147 | havre |
| 12 | 0.603 | logement | 32 | 0.385 | maisonnette | 52 | 0.254 | bas-lieu | 72 | 0.171 | grand_ensemble | 92 | 0.147 | lapinière |
| 13 | 0.596 | maisonnée | 33 | 0.382 | ménil | 53 | 0.249 | manoir | 73 | 0.171 | lieu | 93 | 0.146 | kiosque |
| 14 | 0.593 | nid | 34 | 0.374 | abri | 54 | 0.244 | hôtel | 74 | 0.171 | famille | 94 | 0.146 | intérieur |
| 15 | 0.576 | résidence | 35 | 0.369 | galetas | 55 | 0.241 | standing | 75 | 0.167 | garde-meuble | 95 | 0.144 | bouverie |
| 16 | 0.572 | toit | 36 | 0.368 | lares | 56 | 0.240 | H.L.M. | 76 | 0.167 | deck-house | 96 | 0.144 | parents |
| 17 | 0.545 | bicoque | 37 | 0.350 | lare | 57 | 0.233 | habitacle | 77 | 0.166 | habitat | 97 | 0.143 | tourelle |
| 18 | 0.537 | bâtisse | 38 | 0.338 | clapier | 58 | 0.223 | retraite | 78 | 0.166 | édifice | 98 | 0.143 | mantelet |
| 19 | 0.532 | cahute | 39 | 0.331 | palais | 59 | 0.211 | carbet | 79 | 0.164 | firme | 99 | 0.142 | hangar |
| 20 | 0.502 | baraque | 40 | 0.314 | train_de_vie | 60 | 0.208 | séjour | 80 | 0.163 | tranchée-abri | 100 | 0.141 | pigeonnier |

### *3.2. Similarity matrices and clusterization*

It is also straightforward to build a similarity matrix (see table 2) and to use such matrices to group terms by clusters, i.e. lists of terms which do not all belong to the same clique, but which are closely related semantically. We use nearest-neighbor clustering in this work.

**Table 2 - Similarity matrix**

| | | | | | | | | | | | | |
|---|---|---|---|---|---|---|---|---|---|---|---|---|
| **chacunière** | 1.000 | | | | | | | | | | | |
| **mesnil** | 0.665 | 1.000 | | | | | | | | | | |
| **train_de_maison** | 0.657 | 0.659 | 1.000 | | | | | | | | | |
| **maisonnée** | 0.546 | 0.554 | 0.560 | 1.000 | | | | | | | | |
| **demeure** | 0.456 | 0.470 | 0.449 | 0.394 | 1.000 | | | | | | | |
| **habitation** | 0.414 | 0.433 | 0.403 | 0.344 | 0.852 | 1.000 | | | | | | |
| **maison** | 0.661 | 0.669 | 0.653 | 0.596 | 0.843 | 0.820 | 1.000 | | | | | |
| **pénates** | 0.481 | 0.496 | 0.481 | 0.417 | 0.837 | 0.605 | 0.762 | 1.000 | | | | |
| **logement** | 0.257 | 0.278 | 0.250 | 0.223 | 0.796 | 0.724 | 0.603 | 0.616 | 1.000 | | | |
| **domicile** | 0.457 | 0.464 | 0.462 | 0.394 | 0.780 | 0.670 | 0.767 | 0.620 | 0.588 | 1.000 | | |
| **logis** | 0.507 | 0.514 | 0.506 | 0.447 | 0.795 | 0.666 | 0.810 | 0.706 | 0.623 | 0.892 | 1.000 | |
| **résidence** | 0.305 | 0.310 | 0.308 | 0.262 | 0.727 | 0.609 | 0.576 | 0.511 | 0.554 | 0.840 | 0.627 | 1.000 |

In table 3, the headers are the members of the original cliques including *maison*, grouped in semantically homogeneous clusters, and the associated lists are terms similar with $\sigma > 0.25$ to the center of mass of their header. Terms in blue are from the original cliques, terms in gray are repeats from a previous cluster, and the others could reasonably be aggregated to their head cluster, especially at similarities above $0.35$.

**Table 3 - Clusters around *maison* and their cohorts**

chacunière, mesnil, train_de_maison, maisonnée, demeure, habitation, pénates, maison, logement, domicile, logis, résidence

| | | | | | |
|---|---|---|---|---|---|
| 0.73 maison | 0.55 chacunière | 0.49 foyer | 0.33 bercail | 0.32 cahute | 0.28 lare |
| 0.70 demeure | 0.55 résidence | 0.47 maisonnée | 0.33 château | 0.32 ménil | 0.28 lares |
| 0.67 logis | 0.55 logement | 0.42 nid | 0.33 gîte | 0.30 asile | 0.26 cabane |
| 0.66 domicile | 0.54 mesnil | 0.42 toit | 0.33 bicoque | 0.30 masure | |
| 0.63 pénates | 0.54 train_de_maison | 0.39 bâtisse | 0.33 cassine | 0.29 baraque | |
| 0.62 habitation | 0.49 home | 0.34 chez-soi | 0.33 gourbi | 0.29 immeuble | |

abri, clapier, gîte, nid, asile, retraite, bercail, toit, foyer, habitacle

| | | | | | |
|---|---|---|---|---|---|
| 0.54 abri | 0.41 clapier | 0.37 caponnière | 0.32 logis | 0.29 pare-éclats | 0.26 soue |
| 0.54 nid | 0.40 foyer | 0.36 repaire | 0.32 refuge | 0.29 domicile | 0.26 kiosque |
| 0.50 toit | 0.38 deck-house | 0.36 garde-meuble | 0.31 bouverie | 0.29 home | 0.26 niche |
| 0.49 gîte | 0.38 retraite | 0.36 plaque_de_blindage | 0.30 lapinière | 0.28 taud | 0.26 abrivent |
| 0.47 bercail | 0.37 tranchée-abri | 0.35 tanière | 0.30 hangar | 0.27 reposée | 0.25 cagna |
| 0.46 asile | 0.37 habitation | 0.35 logement | 0.30 mantelet | 0.27 reposoir | 0.25 porcherie |
| 0.43 maison | 0.37 gabionnade | 0.33 havre | 0.30 étable | 0.27 cachette | 0.25 tourelle |
| 0.43 demeure | 0.37 garde-meubles | 0.33 habitacle | 0.29 lieu_sûr | 0.27 auvent | 0.25 brise-vent |
| 0.41 pénates | 0.37 cache | | 0.29 tenderolle | 0.27 darse | |

baraque, bicoque, cahute, cabane, hutte, gourbi, masure, case, cassine, chaumière, maisonnette

| | | | | | |
|---|---|---|---|---|---|
| 0.69 cabane | 0.59 hutte | 0.45 chaumière | 0.37 habitation | 0.30 buron | 0.25 chacunière |
| 0.65 baraque | 0.50 masure | 0.43 maison | 0.36 chaumine | 0.27 train_de_maison | 0.25 appentis |
| 0.64 bicoque | 0.50 maisonnette | 0.42 case | 0.35 carbet | 0.26 chalet | |
| 0.60 cahute | 0.50 gourbi | 0.40 cassine | 0.32 cagna | 0.26 demeure | |

bas-lieu, naissance, origine, descendance, famille, lignée, race, parents, chez-soi, home, intérieur, ménil, lare, lares, ménage, standing, train_de_vie

| | | | |
|---|---|---|---|
| 0.33 famille | 0.31 descendance | 0.27 maison | 0.25 bas-lieu |
| 0.33 race | 0.31 lignée | 0.26 filiation | |

appartement, bouge, taudis, galetas, chalet, pavillon, villa, château, manoir, palais, réduit

| | | | | | |
|---|---|---|---|---|---|
| 0.33 château | 0.32 pavillon | 0.30 galetas | 0.29 maison | 0.27 manoir | 0.26 bouge |
| 0.33 habitation | 0.30 taudis | 0.29 villa | 0.29 chartreuse | 0.27 chalet | |

building, édifice, bâtiment, immeuble, construction, bâtisse, hôtel, campagne, propriété, ferme

| | | | |
|---|---|---|---|
| 0.43 building | 0.40 bâtiment | 0.34 bâtisse | 0.30 H.L.M. |
| 0.41 immeuble | 0.36 édifice | 0.30 construction | |

boîte, entreprise, firme, établissement, prison, commerce, temple, institut, institution, branche, couvert, domesticité, serviteur, domestique, gens, monde, suite

clinique, hôpital, nom, couronne, trône, pigeonnier, lieu, place, séjour, feu

### *3.3. Orthogonalization*

Things get more complicated when two homonyms are semantically disjoint, as is the case with *le barde* and *la barde*:

**Table 4 - First 100 neighbors of *barde***

| From 1 to 20 | | | From 21 to 40 | | | From 41 to 60 | | | From 61 to 80 | | | From 81 to 100 | | |
|---|---|---|---|---|---|---|---|---|---|---|---|---|---|---|
| 1 | 1.000 | barde | 21 | 0.298 | versificateur | 41 | 0.167 | croque-notes | 61 | 0.097 | victimaire | 81 | 0.080 | injurié |
| 2 | 0.839 | aède | 22 | 0.297 | mâche-laurier | 42 | 0.157 | choriste | 62 | 0.096 | flamine | 82 | 0.079 | septemvir |
| 3 | 0.717 | tranche_de_lard | 23 | 0.290 | héros_du_Pinde | 43 | 0.148 | harnais | 63 | 0.096 | prestolet | 83 | 0.079 | lama |
| 4 | 0.625 | chantre | 24 | 0.290 | favori_des_Muses | 44 | 0.147 | prêtre | 64 | 0.094 | iman | 84 | 0.078 | salien |
| 5 | 0.533 | poète | 25 | 0.289 | amant_du_Parnasse | 45 | 0.146 | cigale | 65 | 0.094 | mufti | 85 | 0.078 | brachyne |
| 6 | 0.521 | chanteur | 26 | 0.286 | favori_du_Parnasse | 46 | 0.124 | coryphée | 66 | 0.094 | ratichon | 86 | 0.077 | ménestrier |
| 7 | 0.503 | rhapsode | 27 | 0.285 | nourrisson_du_Parnasse | 47 | 0.119 | trouveur | 67 | 0.093 | ovate | 87 | 0.076 | curé |
| 8 | 0.493 | bardit | 28 | 0.285 | poétereau | 48 | 0.114 | corybante | 68 | 0.093 | utopiste | 88 | 0.076 | talapoin |
| 9 | 0.468 | trouvère | 29 | 0.284 | métromane | 49 | 0.110 | luperque | 69 | 0.091 | ministre_du_culte | 89 | 0.075 | épulon |
| 10 | 0.450 | scalde | 30 | 0.284 | citharède | 50 | 0.110 | muezzin | 70 | 0.090 | pope | 90 | 0.074 | mettre_dans_le_même_sac |
| 11 | 0.427 | troubadour | 31 | 0.282 | crooner | 51 | 0.109 | druide | 71 | 0.090 | mystagogue | 91 | 0.074 | immodérément |
| 12 | 0.421 | minnesinger | 32 | 0.276 | félibre | 52 | 0.105 | eubage | 72 | 0.090 | cantatrice | 92 | 0.074 | sacrificateur |
| 13 | 0.328 | nourrisson_du_Pinde | 33 | 0.274 | duettiste | 53 | 0.105 | parolier | 73 | 0.089 | archiprêtre | 93 | 0.074 | bombardier |
| 14 | 0.323 | amant_des_Muses | 34 | 0.267 | rapsode | 54 | 0.103 | quindecemvir | 74 | 0.086 | sous-ventrière | 94 | 0.073 | lévite |
| 15 | 0.315 | lamelle | 35 | 0.256 | ménestrel | 55 | 0.103 | mollah | 75 | 0.085 | abbé | 95 | 0.073 | englober |
| 16 | 0.313 | favori_d'Apollon | 36 | 0.240 | rimeur | 56 | 0.103 | padre | 76 | 0.084 | curète | 96 | 0.072 | chiennerie |
| 17 | 0.313 | fils_d'Apollon | 37 | 0.233 | rimailleur | 57 | 0.102 | hiérogrammate | 77 | 0.084 | papas | 97 | 0.072 | rabbin |
| 18 | 0.308 | nourrisson_des_Muses | 38 | 0.216 | panne | 58 | 0.100 | saronide | 78 | 0.082 | avarice | 98 | 0.072 | passivité |
| 19 | 0.302 | enfant_d'Apollon | 39 | 0.207 | choreute | 59 | 0.097 | chansonnier | 79 | 0.081 | quindécemvir | 99 | 0.072 | capelan |
| 20 | 0.300 | maître_du_Pinde | 40 | 0.176 | vocaliste | 60 | 0.097 | chapelain | 80 | 0.080 | officiant | 100 | 0.072 | eschatologique |

If we meant *barde* as *aède*, the third neighbor, *tranche_de_lard* is clearly not appropriate, and conversely.

However, in a Euclidean space, the Schmidt orthogonalization procedure does remove this kind of interference. Since term vectors are normalized to unity, one needs simply to subtract from the vector $|\text{barde}\rangle$ the collinear component of the vector $|\text{tranche_de_lard}\rangle$:

$$|\text{barde}\rangle_{\perp tranche_de_lard} = |\text{barde}\rangle - \langle\text{barde}|\text{tranche_de_lard}\rangle \times |\text{tranche_de_lard}\rangle \tag{8}$$

with the following result, where the perturbation due to *tranche_de_lard* is totally eliminated:

**Table 5 - First 100 neighbors of *barde* orthogonalized w.r.t. *tranche_de_lard***

| From 1 to 20 | | | From 21 to 40 | | | From 41 to 60 | | | From 61 to 80 | | | From 81 to 100 | | |
|---|---|---|---|---|---|---|---|---|---|---|---|---|---|---|
| 1 | 0.744 | aède | 21 | 0.396 | poétereau | 41 | 0.217 | croque-notes | 61 | 0.162 | harnais | 81 | 0.130 | lama |
| 2 | 0.697 | barde | 22 | 0.396 | maître_du_Pinde | 42 | 0.215 | lamelle | 62 | 0.161 | prestolet | 82 | 0.130 | papas |
| 3 | 0.634 | chantre | 23 | 0.394 | mâche-laurier | 43 | 0.205 | coryphée | 63 | 0.161 | curète | 83 | 0.126 | soliste |
| 4 | 0.634 | scalde | 24 | 0.393 | favori_des_Muses | 44 | 0.188 | cantatrice | 64 | 0.157 | ministre_du_culte | 84 | 0.124 | talapoin |
| 5 | 0.629 | poète | 25 | 0.389 | nourrisson_du_Parnasse | 45 | 0.185 | quindecemvir | 65 | 0.157 | chapelain | 85 | 0.121 | directeur_de_conscience |
| 6 | 0.622 | chanteur | 26 | 0.389 | amant_du_Parnasse | 46 | 0.183 | mystagogue | 66 | 0.157 | eubage | 86 | 0.120 | utopiste |
| 7 | 0.582 | minnesinger | 27 | 0.387 | héros_du_Pinde | 47 | 0.178 | luperque | 67 | 0.156 | mufti | 87 | 0.120 | prêtraille |
| 8 | 0.548 | trouvère | 28 | 0.380 | félibre | 48 | 0.177 | padre | 68 | 0.155 | saronide | 88 | 0.115 | ceinture_de_sécurité |
| 9 | 0.525 | rhapsode | 29 | 0.380 | rapsode | 49 | 0.177 | mollah | 69 | 0.153 | trouveur | 89 | 0.113 | curé |
| 10 | 0.523 | troubadour | 30 | 0.364 | versificateur | 50 | 0.175 | muezzin | 70 | 0.152 | ovate | 90 | 0.113 | sacrificateur |
| 11 | 0.429 | nourrisson_du_Pinde | 31 | 0.337 | ménestrel | 51 | 0.173 | hiérogrammate | 71 | 0.144 | parolier | 91 | 0.109 | rabbin |
| 12 | 0.415 | crooner | 32 | 0.333 | métromane | 52 | 0.172 | ratichon | 72 | 0.140 | chansonnier | 92 | 0.106 | salien |
| 13 | 0.414 | amant_des_Muses | 33 | 0.319 | choreute | 53 | 0.172 | quindécemvir | 73 | 0.139 | hiérophante | 93 | 0.105 | aumônier |
| 14 | 0.414 | nourrisson_des_Muses | 34 | 0.292 | rimeur | 54 | 0.171 | corybante | 74 | 0.139 | pope | 94 | 0.105 | sous-ventrière |
| 15 | 0.409 | favori_d'Apollon | 35 | 0.287 | rimailleur | 55 | 0.170 | archiprêtre | 75 | 0.137 | diva | 95 | 0.105 | capelan |
| 16 | 0.408 | citharède | 36 | 0.282 | bardit | 56 | 0.169 | druide | 76 | 0.137 | épulon | 96 | 0.105 | affublement |
| 17 | 0.405 | duettiste | 37 | 0.242 | choriste | 57 | 0.168 | victimaire | 77 | 0.135 | abbé | 97 | 0.104 | virtuose |
| 18 | 0.404 | favori_du_Parnasse | 38 | 0.231 | vocaliste | 58 | 0.168 | cigale | 78 | 0.134 | septemvir | 98 | 0.104 | ménestrier |
| 19 | 0.400 | fils_d'Apollon | 39 | 0.224 | panne | 59 | 0.165 | flamine | 79 | 0.133 | musicien | 99 | 0.102 | exécutant |
| 20 | 0.397 | enfant_d'Apollon | 40 | 0.218 | prêtre | 60 | 0.165 | iman | 80 | 0.131 | officiant | 100 | 0.096 | ténor |

The number of terms which can be subtracted is only limited by the noise.

The corresponding clusters associated with $|\text{barde}\rangle_{\perp tranche_de_lard}$ now are:

| **Table 6 - Clusters around *barde* and their cohorts (orthogonalized w.r.t. *tranche_de_lard*)** | | | | | |
|---|---|---|---|---|---|
| aède, barde, poète, chanteur, chantre | | | | | |
| 0.71 chanteur | 0.55 troubadour | 0.44 fils_d'Apollon | 0.43 | 0.42 favori_des_Muses | 0.36 rapsode |
| 0.71 aède | 0.52 barde | 0.44 maître_du_Pinde | amant_du_Parnasse | 0.42 héros_du_Pinde | 0.35 choriste |
| 0.70 scalde | 0.48 rhapsode | 0.43 | 0.42 | 0.42 choreute | 0.34 métromane |
| 0.70 chantre | 0.47 duettiste | nourrisson_du_Pinde | nourrisson_des_Muses | 0.41 amant_des_Muses | 0.31 rimailleur |
| 0.66 poète | 0.46 crooner | 0.43 mâche-laurier | 0.42 enfant_d'Apollon | 0.40 félibre | 0.30 rimeur |
| 0.61 minnesinger | 0.46 citharède | 0.43 | 0.42 favori_d'Apollon | 0.37 versificateur | 0.28 vocaliste |
| 0.55 trouvère | 0.44 poétereau | nourrisson_du_Parnasse | 0.42 favori_du_Parnasse | 0.36 ménestrel | 0.27 coryphée |
| bardit, harnais, prêtre, lamelle, panne, rhapsode, troubadour, trouvère | | | | | |
| 0.29 trouvère | 0.28 minnesinger | | | | |
| 0.29 troubadour | | | | | |

to be compared to the non-orthogonalized result:

| **Table 7 - Clusters around *barde* and their cohorts** | | | | | |
|---|---|---|---|---|---|
| aède, barde, chanteur, chantre, poète | | | | | |
| 0.79 aède | 0.54 trouvère | 0.40 fils_d'Apollon | 0.40 | 0.39 favori_d'Apollon | 0.32 choriste |
| 0.72 chantre | 0.52 troubadour | 0.40 maître_du_Pinde | nourrisson_du_Parnasse | 0.39 choreute | 0.32 tranche_de_lard |
| 0.70 barde | 0.50 rhapsode | 0.40 mâche-laurier | 0.40 | 0.38 favori_du_Parnasse | 0.32 rapsode |
| 0.69 chanteur | 0.42 duettiste | 0.40 | nourrisson_des_Muses | 0.38 amant_des_Muses | 0.32 métromane |
| 0.64 scalde | 0.42 citharède | nourrisson_du_Pinde | 0.39 héros_du_Pinde | 0.38 félibre | 0.28 rimailleur |
| 0.63 poète | 0.42 crooner | 0.40 | 0.39 enfant_d'Apollon | 0.34 versificateur | 0.27 rimeur |
| 0.56 minnesinger | 0.41 poétereau | amant_du_Parnasse | 0.39 favori_des_Muses | 0.33 ménestrel | 0.25 vocaliste |
| bardit, tranche_de_lard, lamelle, rhapsode, troubadour, trouvère | | | | | |
| 0.48 barde | 0.41 aède | 0.34 rhapsode | 0.31 chanteur | 0.28 bardit | |
| 0.42 trouvère | 0.37 minnesinger | 0.33 tranche_de_lard | 0.30 félibre | 0.25 scalde | |
| 0.41 troubadour | 0.35 poète | 0.32 ménestrel | 0.29 chantre | | |
| harnais, panne, prêtre | | | | | |
| 0.35 panne | 0.27 corybante | 0.27 flamine | 0.27 quindecemvir | 0.26 mystagogue | ceinture_de_sécurité |
| 0.34 harnais | 0.27 mufti | 0.27 curète | 0.27 mollah | 0.26 chapelain | 0.26 affublement |
| 0.34 prêtre | 0.27 iman | 0.27 hiérogrammate | 0.27 padre | 0.26 quindécemvir | 0.25 harnois |
| 0.28 muezzin | 0.27 ovate | 0.27 archiprêtre | 0.27 luperque | 0.26 | |

# Conclusion and future work

We have shown how to construct a Euclidean semantic space from a French synonym dictionary by combining random vectors and random projection with the usual vector space model. The process is extremely fast and introduces an amount of noise acceptable in most situations. Updating the semantic space with new synonyms involves handling only a limited number of terms and is thus easily done in real time. As usual, the inner product between two normalized term vectors in the resulting semantic space is interpreted as the similarity between the terms. In additions, the Schmidt orthogonalization process is immediately applicable and can be used to separate semantically disjoint homonyms. The resulting semantic space is directly suited to the generation of practical lists of synonyms and to applications such as word selection for automatic text generation.

This process can be applied to any language, including English, where for example WordNet could be used as a basis. It can also be easily extended to collocations or to contexonyms [15], [16], [17]. In future work, we plan on expanding this method by building a dynamic semantic space from Wikipedia and its “history” pages, whereby the evolution of the meaning of terms as a function of time can be viewed through the evolution of their semantic neighborhoods. This could also be done using time series from newspapers, from radio and television. We are also considering building a real-time semantic space from news and social networking services such as Twitter.